\documentclass[runningheads]{llncs}
\usepackage{graphicx}

\begin{document}
%
\title{GelSplitter: Tactile Reconstruction from Near Infrared and Visible Images}
%
%
\author{Yuankai Lin\inst{1}\orcidID{0000-0001-5764-514X} \and
Yulin Zhou\inst{1}\orcidID{0009-0008-8302-2606} \and
Kaiji Huang\inst{1}\orcidID{0000-0002-0090-3088} \and
Qi Zhong\inst{3} \and
Tao Cheng\inst{2} \and
Hua Yang\inst{1}\orcidID{0000-0002-5430-5630} \and
Zhouping Yin\inst{1}\orcidID{0000-0001-5766-2337}}
\authorrunning{F. Author et al.}
%
\institute{The State Key Laboratory of Digital Manufacturing Equipment and Technology, School of Mechanical Science and Engineering, Huazhong University of Science and Technology, Wuhan 430074, China\\ \email{huayang@hust.edu.cn} \and
College of Urban Transportation and Logistics, Shenzhen Technology University, Shenzhen 518118, China \and
School of Science and Engineering, Tsinghua University, Beijing 100084, China}
\maketitle              
\begin{abstract}
The GelSight-like visual tactile (VT) sensor has gained popularity as a high-resolution tactile sensing technology for robots, capable of measuring touch geometry using a single RGB camera. However, the development of multi-modal perception for VT sensors remains a challenge, limited by the mono camera. 
In this paper, we propose the GelSplitter, a new framework approach the multi-modal VT sensor with synchronized multi-modal cameras and resemble a more human-like tactile receptor.
Furthermore, we focus on 3D tactile reconstruction and implement a compact sensor structure that maintains a comparable size to state-of-the-art VT sensors, even with the addition of a prism and a near infrared (NIR) camera. We also design a photometric fusion stereo neural network (PFSNN), which estimates surface normals of objects and reconstructs touch geometry from both infrared and visible images. 
Our results demonstrate that the accuracy of RGB and NIR fusion is higher than that of RGB images alone. Additionally, our GelSplitter framework allows for a flexible configuration of different camera sensor combinations, such as RGB and thermal imaging.

\end{abstract}

\begin{figure*}[t]
\centering
\includegraphics[width=1.0\textwidth]{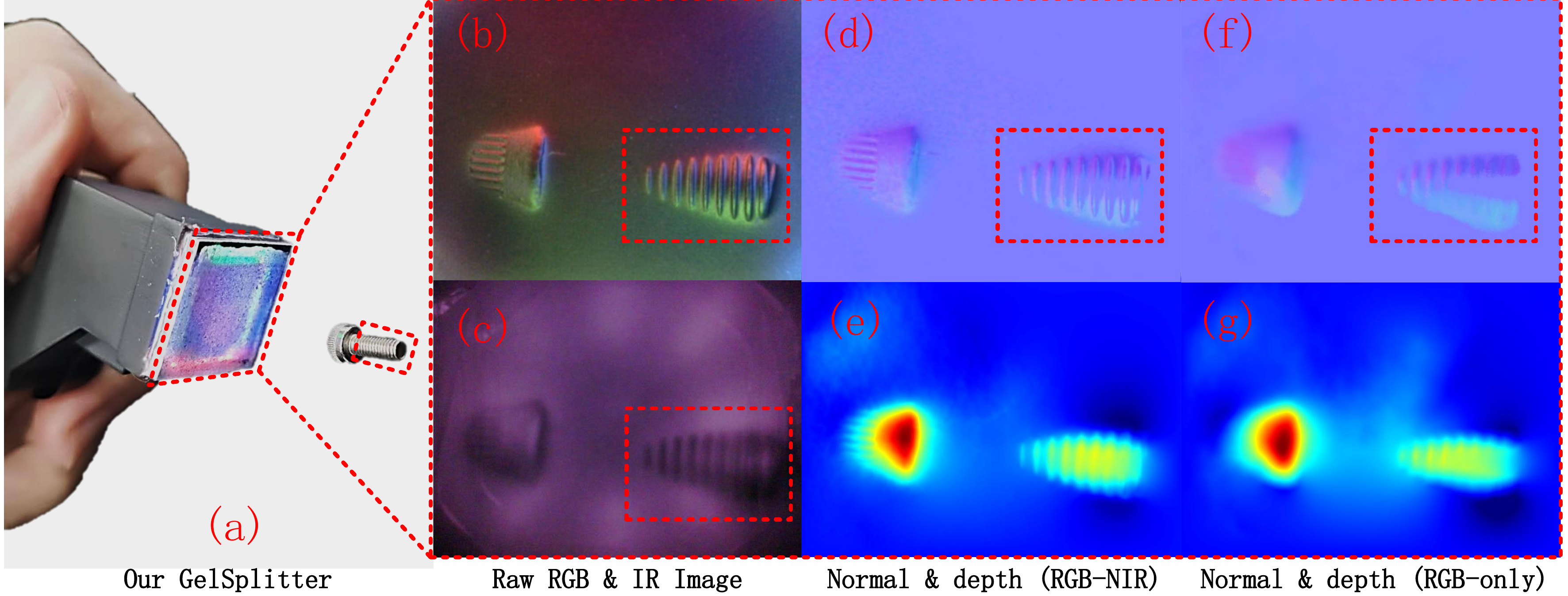} 
\caption{Comparison with RGB-Only method, our GelSplitter offers a framework for the implementation of multi-modal VT sensors,
which can further enhance the performance of tactile reconstruction. (a): Our proposed GelSplitter. (b): A RGB image of our sensor. (c): A corresponding NIR image. (d): The normal map of our PFSNN estimated from the RGB and NIR images. (e): The Depth map reconstructed from the normal map (d). (f): The normal map of the look-up table (LUT) method~\cite{s17122762} estimated from the RGB image. (g): The Depth map reconstructed from the normal map (f). }
\label{fig1}
\end{figure*}
\section{Introduction}
Tactile perception is an essential aspect of robotic interaction with the natural environment~\cite{9028163}. As a direct means for robots to perceive and respond to physical stimuli, it enables them to perform a wide range of tasks and to interact with humans and other objects, such as touch detection~\cite{CastaoAmoros2021TouchDW}, force estimation~\cite{doi:10.1089/soro.2022.0014}, robotic grasping~\cite{9252140} and gait planning~\cite{8823987}. While the human skin can easily translate the geometry of a contacted object into nerve impulses through tactile receptors, robots face challenges in achieving the same level of tactile sensitivity, especially in multi-modal tactile perception.

Visual tactile (VT) sensors such as the GelSights~\cite{s17122762,9811832,9464700} are haptic sensing technology becoming popularity with an emphasis on dense, accurate, and high-resolution measurements, using a single RGB camera to measure touch geometry. However, the RGB-only mode of image sensor restricts the multi-modal perception of VT sensor. RGB and near infrared (NIR) image fusion is a hot topic for image enhancement, using IR images to enhance the RGB image detail and improve measurement accuracy~\cite{9055063,ZHAO2020107734}. From this perspective, a multi-modal image fusion VT sensor is promoted to enrich the tactile senses of the robot.

In this paper, we present the GelSplitter with RGB-NIR fusion as a solution of the above challenge. This novel design integrates a splitting prism to reconstruct tactile geometry from both NIR and RGB light cameras. Our GelSplitter offer a framework for the implementation of multi-modal VT sensors, which can further enhance the tactile capabilities of robots, as shown in Fig. \ref{fig1}. Additionally, our GelSplitter framework allows for a flexible configuration of different camera sensor combinations, such as RGB and thermal imaging. In summary, our contribution lies in three aspects:
\begin{itemize}
\item 
We propose the GelSplitter, a new framework to design multi-modal VT sensor. 
\item 
Based on the framework, we focus on the task of 3D tactile reconstruction and fabricate a compact sensor structure that maintains a comparable size to state-of-the-art VT sensors, even with the addition of a prism and camera. 
\item 
A  photometric fusion stereo neural network (PFSNN) is implemented to estimate surface normals of objects and reconstructs touch geometry from both infrared and visible images. The common issue of data alignment between the RGB and IR cameras is addressed by incorporating splitting imaging.
Our results demonstrate that our method outperforms the RGB-only VT sensation.
\end{itemize}

The rest of the paper is organized as follows: In Sect. 2, we briefly introduce some related works of VT sensors and multi-modal image fusion. Then, we describe the main design of the GelSplitter and FPSNN in Sect. 3. In order to verify the performance of our proposed method, experimental results are presented in Sect. 4.
\begin{figure*}[t]
\centering
\includegraphics[width=1.0\textwidth]{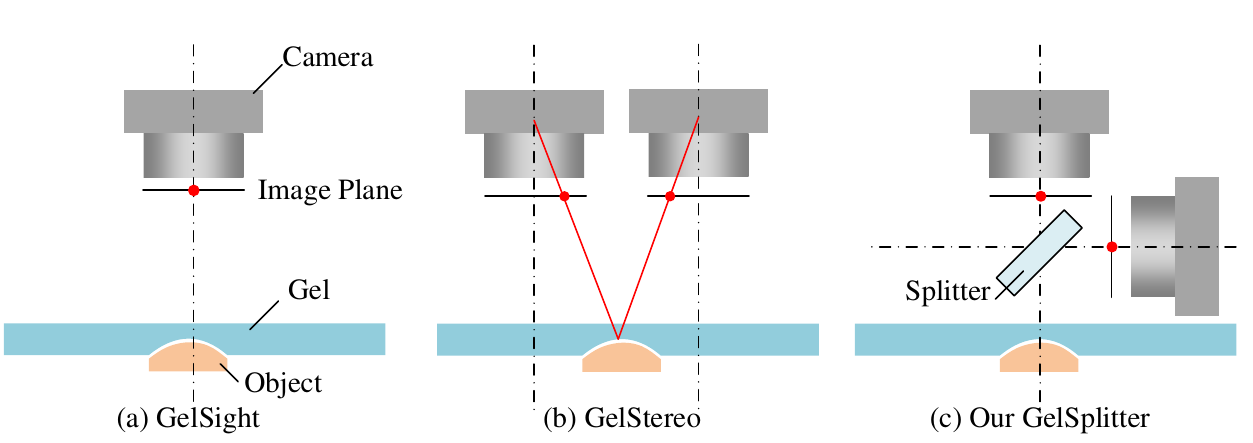} 
\caption{Comparison with the existed VT sensors, our GelSplitter extends the dimensionality of perception while maintaining the same image plane. (a): The Imaging mode of a typical Geisight-like~\cite{s17122762} VT sensor. (b): The Imaging mode of a GelStereo~\cite{10035880} VT sensor. (c): Our imaging mode of GelSplitter maintains the optical centres of the different cameras and align the multi-modal data. In addition, our framework allows for a flexible configuration of different camera sensor combinations, such as RGB and thermal imaging.}
\label{fig2}
\end{figure*}
\section{Related Work}
\subsection{VT Sensor}
GelSight~\cite{s17122762} is a well-established VT sensor that operates like a pliable mirror, transforming physical contact or pressure distribution on its reflective surface into tactile images that can be captured by a single camera. Building upon the framework of the GelSight, various VT sensors~\cite{8593661,9811832,9560783} are designed to meet diverse application requirements, such as ~\cite{8794113,9762175,10122053,9772441}. Inspired by GelSight, DIGIT~\cite{9018215} improves upon past the VT sensors by miniaturizing the form factor to be mountable on multi-fingered hands. 

Additionally, there are studies that explore the materials and patterns of reflective surfaces to gather other modal information of touch sensing. For example, FingerVision~\cite{doi:10.1142/S0219843619400024} provides a multi-modal sensation with a completely transparent skin. HaptiTemp~\cite{9652522} uses thermochromic pigments color blue, orange, and black with a threshold of 31$^{\circ}$C, 43$^{\circ}$C, and 50$^{\circ}$C, respectively on the gel material, to enable high-resolution temperature measurements. DelTact~\cite{9849124} adopts an improved dense random color pattern to achieve high accuracy of contact deformation tracking.

The VT sensors mentioned above are based on a single RGB camera. Inspired by binocular stereo vision, GelStereo~\cite{9464700} uses two RGB cameras to achieve tactile geometry measurements. Further, the gelstereo is developed in six-axis force/torque estimation~\cite{9981100} and geometry measurement~\cite{9654189,10035880}.

In current research, one type of camera is commonly used, with a small size and a simplified optical path. The RGB camera solution for binocular stereo vision effectively estimates disparity by triangulation. However, data alignment is a common issue to different kinds of camera images~\cite{9157219}, because multi-modal sensors naturally have different extrinsic parameters between modalities, such as lens parameters and relative position. In this paper, two identical imaging windows are fabricated by a splitting prism, which filter RGB component (bandpass filtering at 650$nm$ wavelength) and IR component (narrowband filtering at 940$nm$ wavelength).

\subsection{Multi-modal Image Fusion}
Multi-modal image fusion is a fundamental task for robot perception, healthcare and autonomous driving~\cite{9094690,SINGH2023104020}. However, due to high raw data noise, low information utilisation and unaligned multi-modal sensors, it is challenging to achieve a good performance.
In these applications, different types of datasets are captured by different sensors such as infrared (IR) and RGB image~\cite{ZHAO2020107734}, computed tomography (CT) and positron emission tomograph (PET) scan~\cite{9031751}, LIDAR point cloud and RGB image~\cite{lin2022dynamic}.

Typically, the fusion of NIR and RGB images enhances the image performance and complements the missing information in the RGB image. DenseFuse~\cite{8580578} proposes an encoding network that is combined with convolutional layers, a fusion layer, and a dense block to get more useful features from source images. DarkVisionNet~\cite{jin2022darkvisionnet} extracts clear structure details in deep multiscale feature space rather than raw input space by a deep structure. 
MIRNet~\cite{9756908} adopts a multi-scale residual block to learn an enriched set of features that combines contextual information from multiple scales, while simultaneously preserving the high-resolution spatial details. 

Data alignment is another important content of multi-modal image fusion. Generally, the targets corresponding to multiple sensors are in different coordinate systems, and the data rates of different sensors are diverse. To effectively utilize heterogeneous information and obtain simultaneous target information, it is essential to map the data onto a unified coordinate system with proper time-space alignment~\cite{8943388,SINGH2023104020}.

Following the above literature, PFSNN is designed for the gelsplitter, completing the missing information of RGB images by NIR image. It generates refined normal vector maps and depth maps. In addition, splitting prisms are embeded to unify the image planes and optical centres of the different cameras and align the multi-modal data.

\begin{figure*}[t]
\centering
\includegraphics[width=1.0\textwidth]{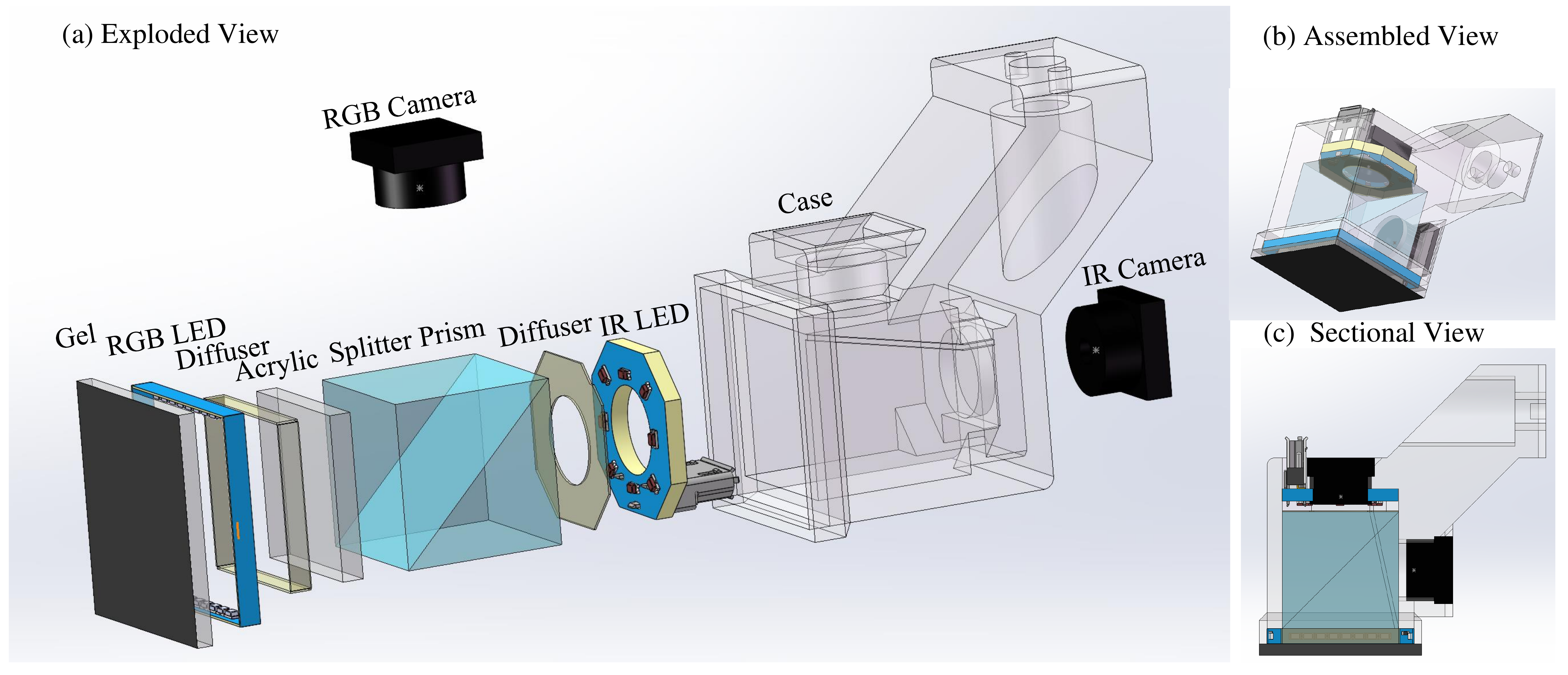} 
\caption{ The design of our GelSplitter, including the exploded view showing inner components, the assembled CAD model, and the Sectional view of the sensor.}
\label{fig3}
\end{figure*}
\section{Design and Fabrication}
Our design aims to achieve high-resolution 3D tactile reconstruction while maintaining a compact shape. Fig. \ref{fig3} shows the components and schematic diagram of the sensor design. Following, we describe the design principles and lessons learned for each sensor component.
\subsection{Prism and Filters}
To capture the NIR image and the RGB image, a splitting prism, band pass filter, narrow band filter and diffuser were prepared 
. These components ensure that the images are globally illuminated and homogeneous.

The splitting prism is a cube with a side length of 15$mm$, as shown in Fig. \ref{fig4} (a). It has a spectral ratio of 1:1 and a refractive index of 1.5168, creating two identical viewports. Among the six directions of the cube, except for two directions of the cameras and one direction of the gel, the other three directions are painted with black to reduce secondary reflections. 

The diffuser in lighting can produce more even global illumination as it distributes light evenly throughout the scene. The octagonal diffuser needs to be optically coupled to the splitting prism to avoid reflection from the air interface, as shown in Fig. \ref{fig4} (b). “3M Diffuser 3635-70” is used for the sensor.

A 650$nm$ bandpass filter and 940$nm$ narrowband filter are used to separate the RGB component from the NIR component. The lens and filter are integrated together, as shown in Fig. \ref{fig4} (c). 

\subsection{Lighting}
\begin{figure*}[t]
\centering
\includegraphics[width=1.0\textwidth]{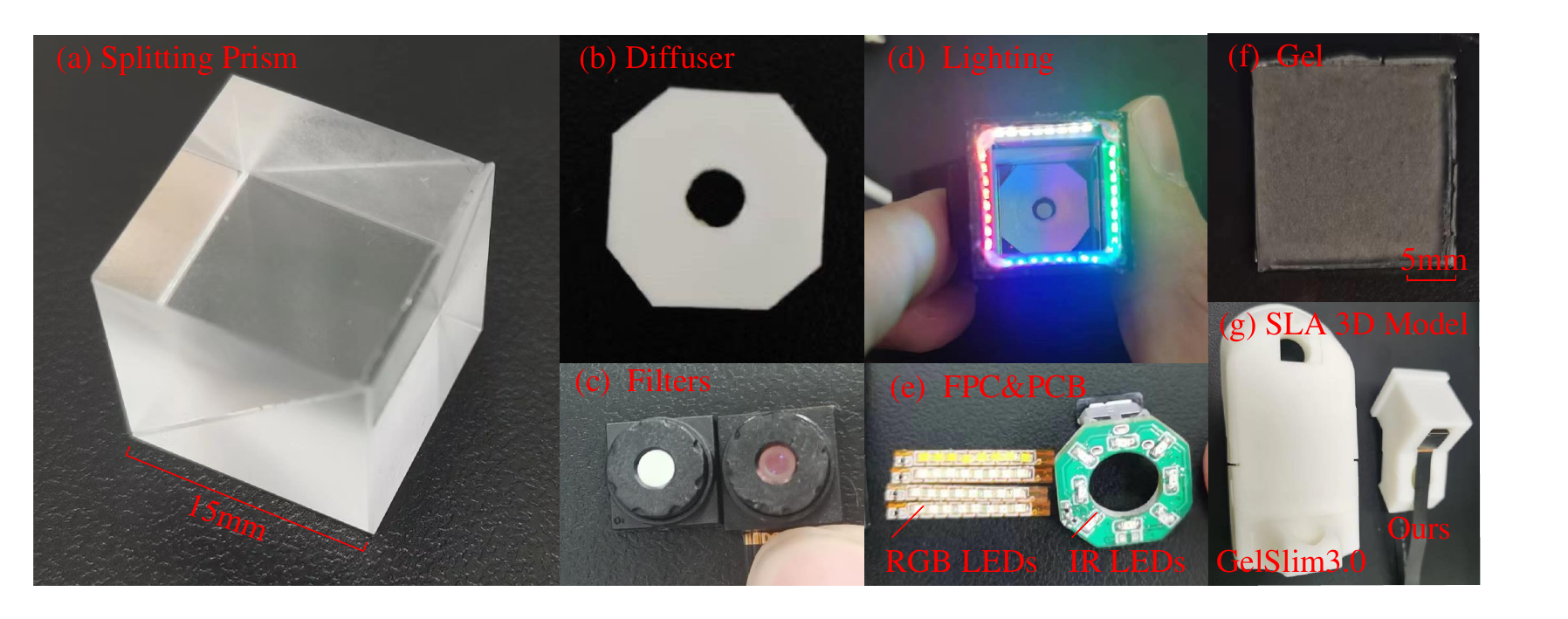} 
\caption{The inner components and detials of our GelSplitter. (a): The splitting prism. (b): The diffuser. (c): The 650$nm$ bandpass filter and 940$nm$ narrowband filter. (d): The lighting. (e): The FPC and PCB to drive LEDs. (f): The Gel with reflective covering. (g): The Silk screen printing plate. }
\label{fig4}
\end{figure*}

In this paper, five types of colour LEDs are provided for illumination in different directions, including red ( Type NCD0603R1, wavelength 615$\sim$630$nm$), green ( Type NCD0603W1, wavelength 515$\sim$530$nm$), blue ( Type NCD0603B1, wavelength 463$\sim$475$nm$), white ( Type NCD0603W1) and infrared ( Type XL-1608IRC940, wavelength 940$nm$), as shown in Fig. \ref{fig4} (d). These LEDs are all in 0603 package size (1.6×0.8 $mm$).

Moreover, in order to represent surface gradient information, LEDs are arranged in two different ways. Firstly, red, green, blue and white LEDs can be arranged in rows, illuminating the four sides of the gel. This allows for a clear representation of surface gradients from different angles. Secondly, infrared LEDs can be arranged in a circular formation above the gel, surrounding the infrared camera. This allows IR light to shine on the shaded areas, supplementing the missing gradient information.

Finally, we designed the FPC and PCB to drive LEDs, as shown in Fig. \ref{fig4} (e). The LED brightness is configured with a resistor, and its consistency is ensured by a luminance meter.

\subsection{Camera}
Both NIR and RGB cameras are used the common CMOS sensor OV5640, manufactured by OmniVision Technologies, Inc. The OV5640 supports a wide range of resolution configurations from 320×240 to 2592×1944, as well as auto exposure control (AEC) and auto white balance (AWB). Following the experimental setting of GelSights~\cite{9028163}, our resolution is configured to 640×480. AEC and AWE are disabled to obtain the linear response characteristics of the images.

To capture clear images, it is essential to adjust the focal length by rotating the lens and ensure that the depth of field is within the appropriate range. 
Although ideally, the optical centres of the two cameras are identical. In practice, however, there is a small assembly error that requires fine-grained data alignment.

Both RGB and NIR cameras are calibrated and corrected for aberrations using the Zhang's calibration method implemented in OpenCV. Random sample consistency (RANSAC) regression is used to align the checkerboard corners of the two cameras' images to achieve data alignment.
\subsection{Elastomer}
Our transparent silicone is coated with diffuse reflective paint. We choose low-cost food grade platinum silicone which operates in a 1:1 ratio. The silicone is poured into a mould and produced as a soft gel mat with a thickness of 1.5 $mm$. In our experiment, we found that a gel with a Shore hardness of 10A is both pliable enough to avoid breakage and capable of producing observable deformations. To achieve defoaming, it is necessary to maintain an environmental temperature of approximately 10°C and apply a vacuum pressure of -0.08 MPa.

\subsection{SLA 3D Model}
We use stereo lithography appearance (SLA) 3D printing to create the case. Compared to fused deposition modelling (FDM) 3D printing of PLA, SLA technology has a higher precision for assemble requirements of the splitting prism. Despite the addition of a prism and a NIR camera, our case still maintains a comparable size to the state-of-the-art VT sensor, GelSlim 3.0~\cite{9811832}, as shown in Fig. ~\ref{fig4} (g). 


\begin{figure*}[t]
\centering
\includegraphics[width=1.0\textwidth]{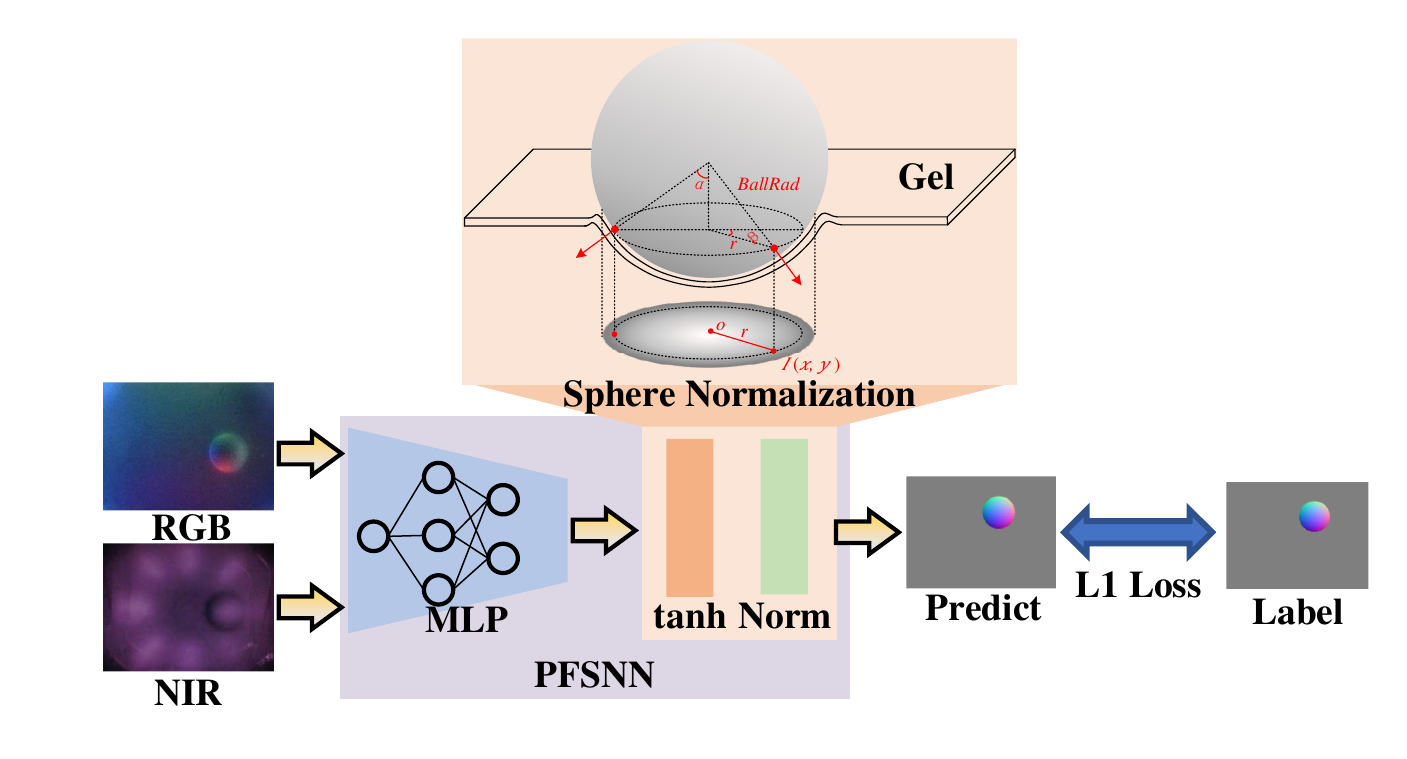} 
\caption{We propose PFSNN to fuse RGB images and NIR images from GelSplitter, and estimate dense normal maps. PFSNN is composed of multi-layer perceptron (MLP) and sphere normalization, and is supervised by L1 loss function. Furthermore, the fast poisson algorithm~\cite{s17122762} can be utilized to solve depth maps based on normal maps.
}
\label{fig5}
\end{figure*}    
\begin{table}[t]
\caption{Network Architecture of PFSNN.}
\label{tab1}
\centering
\begin{tabular}{ccccc}
\hline
Layer & Operator     & Kernel Size & Input channels           & Output Channels \\ \hline
1     & Concat       & -           & 4(RGB-NIR)+4(Background) & 8               \\
2     & Conv2d       & 1           & 8                        & 128             \\
3     & Relu         & -           & 128                      & 128             \\
4     & Conv2d       & 1           & 128                      & 64              \\
5     & Relu         & -           & 64                       & 64              \\
6     & Conv2d       & 1           & 64                       & 3               \\
7     & Relu         & -           & 3                        & 3               \\
8     & Tanh         & -           & 3                        & 3               \\
9     & Normalize    & -           & 3                        & 3               \\
10    & $x=0.5x+0.5$ & -           & 3                        & 3               \\ \hline
\end{tabular}
\end{table}
\section{Measuring 3D Geometry}
In this section, We introduce PFSNN which fuses RGB images and NIR images from GelSplitter, and estimate dense normal maps. Following, we describe the components and implementation details of PFSNN. 
\subsection{PFSNN}
\subsubsection{Network Architecture} of the PFSNN is shown in the table \ref{tab1}. It is composed of multi-layer perceptron (MLP) and sphere normalization. Compared to the look-up table (LUT) method~\cite{s17122762}, the MLP network is trainable. Through its non-linear fitting capability, the algorithm can combine and integrate information from both RGB and NIR sources. 
\subsubsection{Sphere Normalization} is derived from a physical model of the spherical press distribution and outputs a unit normal vector map, as shown in Fig. \ref{fig5}. It is defined as:
 \begin{equation} 
n=\frac{tanh(x)}{max(\left \| tanh(x) \right \|_2,\epsilon  )} ,
\end{equation}
where $x$ is the output of the MLP, and $\epsilon$ is a small value ($10^{-12}$ in this paper) to avoid division by zero.  
Furthermore, the fast poisson algorithm [28] can be utilized to solve depth maps based on normal maps. It is defined as:
 \begin{equation} 
d=Fast\_Poisson(n).
\end{equation}
\subsubsection{Implementation Details.} 
\begin{figure*}[t]
\centering
\includegraphics[width=1.0\textwidth]{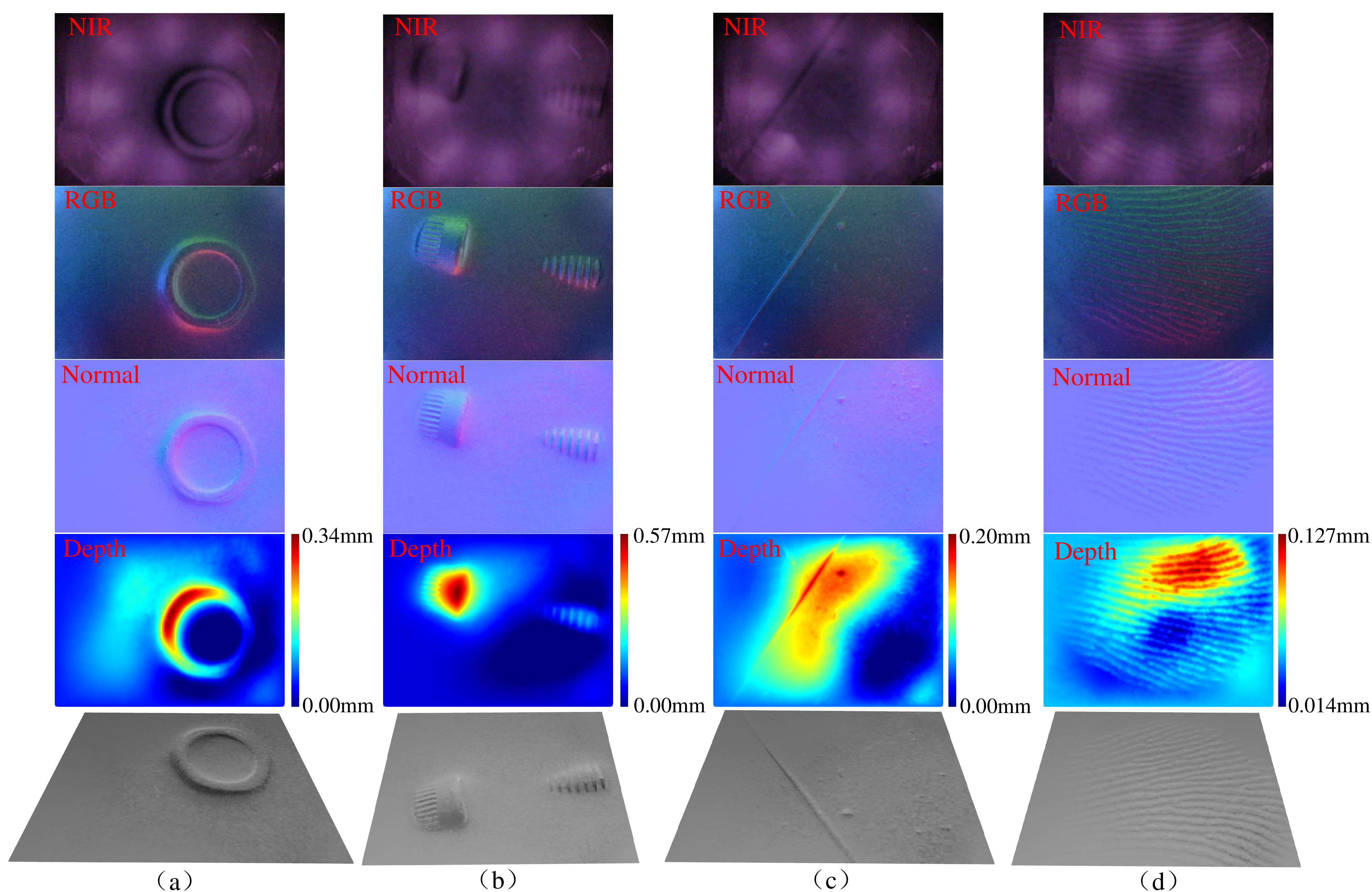} 
\caption{Result of 3D tactile reconstruction from RGB-NIR. (a):A screw cap. (b):A screw. (c): A hint of hair. (d): A fingerprint. Even though GelSplitter is used to touch these items for the first time, remarkably clear shapes are still able to be reconstructed through the 3D tactile reconstruction process. }
\label{fig6}
\end{figure*}

PFSNN requires a small amount of data for training. In this paper, only five images of spherical presses were collected, four for training and one for validation. We test the model with screw caps, screws, hair and fingerprints that the PFSNN never seen before, as shown in Fig. \ref{fig6}.

All of our experiments are executed on a NVIDIA RTX 3070 laptop GPU. Our method is implemented in the PyTorch 2.0 framework,  trained with an ADAM optimizer. The batch size is set to 64. The learning rate is set to $0.01$ for  20 epochs. There is no use of data enhancement methods.

\begin{table}[t]
\caption{Experiment Result of PFSNN.}
\label{tab2}
\centering
\begin{tabular}{ccccc}
\hline
         & LUT w/o NIR & LUT w. NIR & PFSNN w/o NIR & PFSNN w. NIR \\ \hline
MAE(°) & 9.292       & 8.731      & 6.057         & 5.682        \\ \hline
\end{tabular}
\end{table}

\subsection{Result of 3D tactile Reconstruction}
The process of 3D tactile reconstruction can be divided into two main steps. The first step involves calculating the normal map from RGB-NIR. The second step is to apply the fast Poisson solver~\cite{s17122762} to integrate the gradients and obtain the depth information. This helps to create a detailed 3D tactile reconstruction, which can be very useful for various applications.

The screw cap, screw, hair, fingerprint is chosen to qualitative testing of resolution and precision, as shown in Fig. \ref{fig6}. In NIR images, the opposite property to RGB images is observed, with higher depth map gradients being darker. This means that our design allows the NIR image to contain information that is complementary to the RGB image and illuminates the shaded parts of the RGB image,as shown in Fig. \ref{fig6} (a).
In addition, our splitting design of imaging allows both the normal vector map and the reconstructed depth map to have a clear texture, as shown in Fig. \ref{fig6} (b), where the threads of the screw are clearly reconstructed.
Hair strands (diameter approx. 0.05$mm$ $\sim$ 0.1$mm$) and fingerprints(diameter approx. 0.01$mm$$\sim$0.02$mm$) are used to test the minimum resolution of the GelSplitter, as shown in Fig. \ref{fig6} (c) (d). 

In addition, our splitter can be easily switched between RGB and RGB-NIR modes and provides a fair comparison of the results, as shown in the Tab. \ref{tab2}. The LUT method~\cite{s17122762} is employed as a baseline to verify the validity of our method. The results show that the addition of NIR reduces the normal error by 0.561° and 0.375° for LUT and PSFNN respectively. Our PFSNN outperforms the LUT, decreasing the error by over 40\%.

\section{Conclusion}
 
In this paper, we proposed a framework named GelSplitter to implement multi-modal VT sensor. 
Furthermore, we focus on 3D tactile reconstruction and designed a compact sensor structure that maintains a comparable size to state-of-the-art VT sensors, even with the addition of a prism and camera. We also implemented the PFSNN to estimate surface normals of objects and reconstruct touch geometry from both NIR and RGB images. 
Our experiment results demonstrated the performance of our proposed method.

\section*{Acknowledgments}
This work was supported by the Guangdong University Engineering Technology Research Center for Precision Components of Intelligent Terminal of Transportation Tools (Project No.2021GCZX002), and Guangdong HUST Industrial Technology Research Institute, Guangdong Provincial Key Laboratory of Digital Manufacturing Equipment.
%
%
%

\bibliographystyle{splncs04}
\bibliography{reference.bib}
%




\end{document}